\pdfoutput=1
\PassOptionsToPackage{table,xcdraw,svgnames}{xcolor}

\documentclass[11pt]{article}

\usepackage{acl}  

\usepackage{times}
\usepackage{latexsym}
\usepackage{amsmath}
\usepackage{booktabs} 
\usepackage{caption}
\usepackage{float}

\usepackage[T1]{fontenc}

\usepackage[utf8]{inputenc}

\usepackage{microtype}

\usepackage{inconsolata}

\usepackage{graphicx}
\usepackage{hyperref}
\hypersetup{
    colorlinks=true,
    linkcolor=[rgb]{0, 0, 0.5},  
    urlcolor=[rgb]{0, 0, 0.5},   
    citecolor=[rgb]{0, 0, 0.5},  
    linkbordercolor={1 1 1},     
    urlbordercolor={1 1 1},      
    citebordercolor={1 1 1}      
}

\title{Waste-Bench: A Comprehensive Benchmark for Evaluating VLLMs in Cluttered Environments}

\author{
  {\bf Muhammad Ali} \and {\bf Salman Khan} \\
  Mohamed Bin Zayed University of Artificial Intelligence\\
  \texttt{muhammad.ali@mbzuai.ac.ae}
}

\begin{document}

\maketitle
\begin{abstract}
 Recent advancements in Large Language Models (LLMs) have paved the way for Vision Large Language Models (VLLMs) capable of performing a wide range of visual understanding tasks. While LLMs have demonstrated impressive performance on standard natural images, their capabilities have not been thoroughly explored in cluttered datasets where there is complex environment having deformed shaped objects. In this work, we introduce a novel dataset specifically designed for waste classification in real-world scenarios, characterized by complex environments and deformed shaped objects. Along with this dataset, we present an in-depth evaluation approach to rigorously assess the robustness and accuracy of VLLMs.
The introduced dataset and comprehensive analysis provide valuable insights into the performance of VLLMs under challenging conditions. Our findings highlight the critical need for further advancements in VLLM's robustness to perform better in complex environments. The dataset and code for our experiments will be made publicly available.
\end{abstract}

\section{Introduction}

In recent years, Large Language Models (LLMs) \cite{chung2024scaling, achiam2023gpt, touvron2023llama} have demonstrated remarkable capabilities in understanding, reasoning, and generating text for a diverse range of open-ended tasks. Models such as PaLM 2 \cite{anil2023palm} and Falcon \cite{refinedweb} have showcased exceptional performance in commonsense reasoning, multilingual applications, and various Natural Language Processing (NLP) tasks. Building on their success,  Vision-Language Large Models (VLLMs) \cite{fang2023eva, touvron2023llama, zheng2023judging} have emerged, extending these capabilities to multimodal domains by integrating visual and textual data. Notable examples, including multimodal GPT-4 and open-source models like LLaVA \cite{achiam2023gpt, liu2023improvedllava, liu2024llavanext}, excel in a variety of multimodal tasks, demonstrating their versatility in real-world applications \cite{hu2023rsgpt, vinyals2015show, 9093452}.

Despite advancements in Vision-Language Models (VLLMs), their application in complex, cluttered environments remains underexplored. Traditional object detectors, such as Faster R-CNN \cite{ren2015faster} and YOLO \cite{redmon2016you}, are effective for visual localization and classification tasks. Traditional models are confined to fixed labels and cannot handle open-ended, context-aware questions.
Vision-language models, by aligning images with text, can answer queries such as “Which items are recyclable under this lighting?” or “How many soft-plastic items overlap metal objects?”, a capability essential for cluttered waste-sorting scenes.
To address these challenges, we propose Waste-Bench, a benchmark designed to evaluate the robustness and reasoning capabilities of VLLMs in the context of waste classification. Unlike existing benchmarks, such as SEED-Bench \cite{li2023seed} and MV-Bench \cite{li2024mvbench}, which focus primarily on general visual comprehension, Waste-Bench targets the unique complexities of real-world waste management scenarios, including cluttered scenes, deformed objects, and ambiguous visual cues. By systematically evaluating pre-trained VLLMs, Waste-Bench highlights their baseline capabilities and limitations, offering actionable insights to guide the improvement of future VLLMs.

Furthermore, Waste-Bench is intended to complement existing datasets, enriching them with challenging scenarios that encourage greater robustness and adaptability in models. By incorporating diverse data distributions into training pipelines, models can achieve better trade-offs between task-specific robustness and generalization. This approach aligns with robust learning paradigms, which suggest that exposure to diverse, challenging data distributions can enhance model generalization while minimizing the risks of performance degradation on simpler tasks \cite{havrilla2024surveying}.
To improve VLLMs in such environments, techniques like domain adaptation and adversarial training \cite{ganin2016dann, sun2019testtime} can be employed to expose the models to more realistic, noisy, and cluttered data. Additionally, incorporating multi-modal learning, including multispectral data, and using data augmentation strategies during training \cite{madry2018pgd} can help VLLMs better adapt to complex, cluttered environments. Fine-tuning models on Waste-Bench’s diverse and complex scenarios ensures that they become more robust to variations in visual cues, allowing them to handle the unique challenges of waste classification tasks effectively.

Models trained on simpler datasets often experience a performance drop when evaluated in cluttered environments, primarily due to insufficient exposure to noise, occlusions, and ambiguities during training.  To address this challenge, Waste-Bench exposes models to more complex and realistic waste classification scenarios. By training models on these challenging conditions, Waste-Bench helps to reduce the performance gap between regular and cluttered environments, improving model generalization without sacrificing accuracy.
Although the performance discrepancy between regular and cluttered environments has not been extensively studied in VLLMs, this issue is well-known in traditional vision tasks. In literature, various waste classification methods have been proposed \cite{xia2024yolo, mao2021recycling, feng2022intelligent, meng2022mobilenet}, they pose limitations in the presence of complex scenarios where there exists an unclear boundary information. Waste-Bench aims to mitigate this gap by training models on more challenging, real-world data, making them more adaptable and robust.
Our contributions are as follows:
\begin{itemize}
 \item A Waste-Bench designed to evaluate the robustness and reasoning capabilities of VLLMs in waste classification, addressing the complexities of real-world applications.
\item  We evaluate VLLMs, uncovering significant challenges, especially in reasoning within cluttered scenes with deformed objects.
\item We identify that VLLMs struggle with various tasks on Waste-Bench, guiding future waste management improvements.
\end {itemize}
\section{Related Work}
\textbf{Vision Large Language Models} 
 (VLLMs) 
\cite{zhu2023minigpt,shao2023tiny} have demonstrated remarkable capabilities in engaging with visual content, offering a wide range of potential applications. Notable models in this domain include Qwen \cite{qwen}, which has consistently demonstrated superior performance across various downstream tasks.  
Gemini-Pro and GPT-4o \cite{reid2024gemini, openai2024gpt4o} exemplifies state-of-the-art performance with its advanced reasoning and interaction capabilities, paving the way for the development of versatile multimodal conversational assistants. All these models perform extremely well on wide range of image understanding tasks like caption generation, visual question answering and so on.
These models accept both visual and textual inputs and generate textual responses. From an architectural perspective, VLLMs typically combine pre-trained vision backbones \cite{fang2023eva} with large language models \cite{touvron2023llama, zheng2023judging} using connector modules such as MLP adapters, Q-former \cite{dai2024instructblip}, and gated attention \cite{alayrac2022flamingo}.\\
\textbf{Benchmarking VLLMs}
With the growing number of VLLMs emerging in the research community, several benchmarks have been proposed to evaluate and quantify these models for benchmarking and analysis purposes. Notable benchmarks in this domain include SEED-Bench \cite{li2023seed}, which evaluates the visual capabilities of both image and video LMMs across multiple dimensions, and MV-Bench \cite{li2024mvbench}, which curates challenging tasks to evaluate the spatial and temporal understanding of VLLMs. While these benchmarks provide effective insights into model performance, they primarily focus on general visual comprehension metrics.
 However, none of them specifically target complex cluttered environments and deformed shaped objects. 
In contrast, Waste-Bench is a comprehensive benchmark designed to assess the robustness and reasoning capabilities of VLLMs in waste classification.

\section{Waste-Bench}
\begin{figure*}
  \centering
   \includegraphics[width=.8\textwidth]{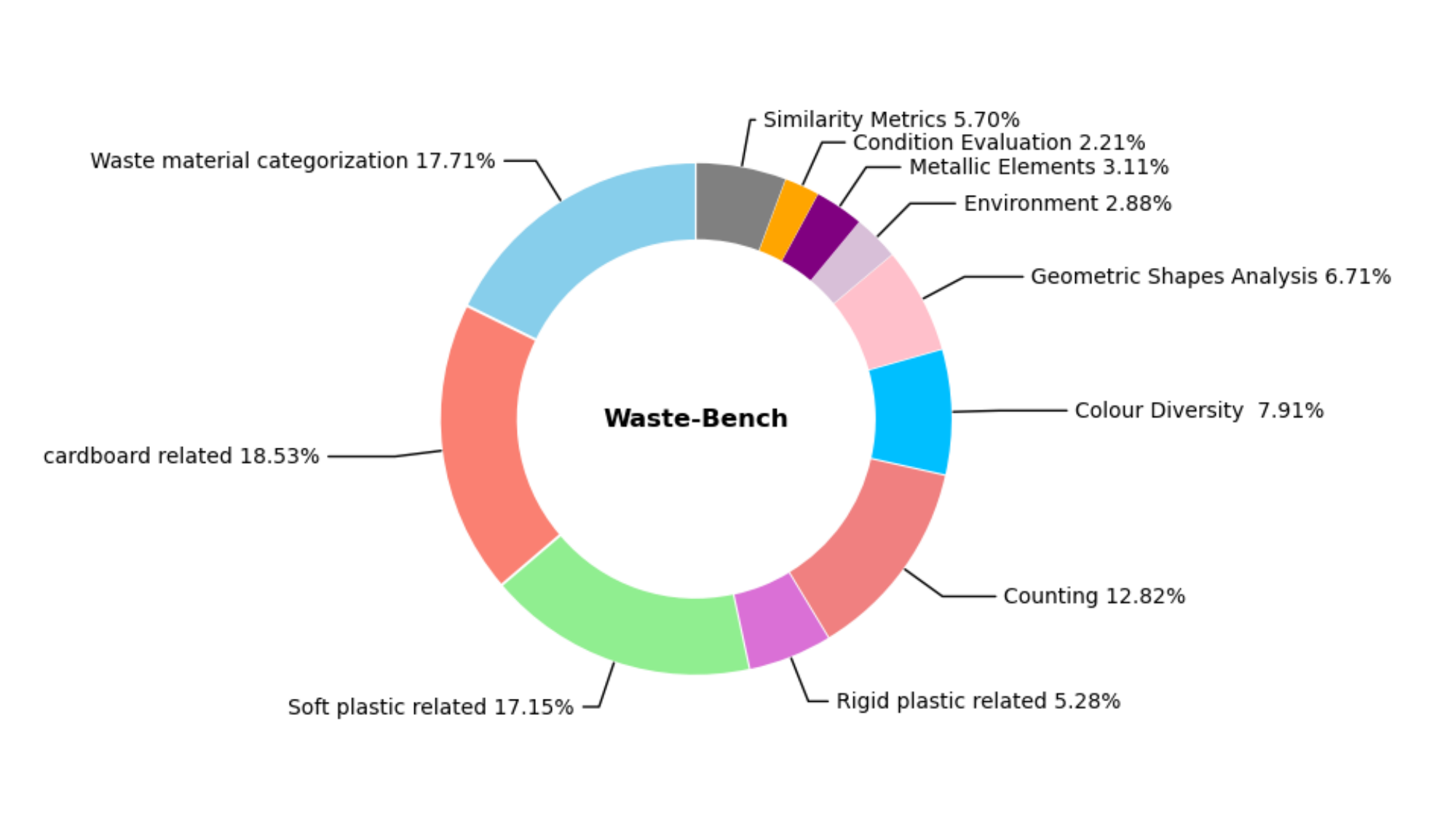}
  \caption{Waste-Bench comprises of 11 diverse complex question categories  encompassing a variety of waste images context.}
  \label{fig:comparison}
\end{figure*}

In this work, our objective is to develop a comprehensive benchmark to evaluate the robustness and reasoning capabilities of VLLMs in various complex and cluttered visual environments, spanning diverse scenarios. To achieve this, we introduce Waste-Bench. Initially, we offer a holistic overview of Waste-Bench and outline the diversity of questions it contains. Following this, we detail the creation process of Waste-Bench in Section 3.2. Performance evaluation including experiments and results are given in Section 4 and 5 respectively.
\subsection{Waste-Bench Dataset}
Waste-Bench encompasses 11 different question categories and 9,520 high-quality open-ended question-answer (QA) pairs, spanning 952 high-quality images with an average of 10 questions per image. These questions cover diverse categories related to real-world waste classification scenarios, including individual classification of waste classes, multi-class classification, shapes of objects, and colors. This comprehensive dataset is designed to rigorously test the capabilities of VLLMs in handling complex and cluttered visual environments. The question types and word cloud of frequent keywords is given in Appendix \ref{frequency}.

\label{headings}
\subsubsection{Waste-Bench Different Question Types}
To assess the robustness and reasoning capabilities of VLLMs in the Waste-Bench benchmark, we ensure it contains various question types to encompass a wide range of real-world complex and cluttered visual environments within each image. Below, we provide a detailed definition of the Waste-Bench as given in Figure \ref{fig:comparison}.
\begin{itemize}
\item Single Class Classification (Cardboard, Metal, Soft Plastic, Rigid Plastic):
This category includes questions that require the model to classify individual waste items into one of the specified single classes. The questions aim to determine whether the model can accurately identify and distinguish between different types of materials commonly found in waste.
\item Multiclass Categorization:
In this category, the models are challenged with images containing multiple deformed waste items that need to be classified into more than one category. The goal is to assess the model's ability to handle complex scenes where multiple waste types are present and need to be accurately categorized.
\item Counting:
This category involves tasks where the model must count the number of specific items or categories within an image. For example, counting the number of cardboard pieces or the number of recyclable items in a cluttered environment. The questions are designed to evaluate the model's precision in quantifying objects in a scene.
\item Color Diversity:
This question type  tests the model's ability to distinguish and identify items based on color. Tasks in this category include identifying objects of a specific color or categorizing items by color diversity. It assesses the model's capability to utilize color as a key feature in classification.
\item Geometric Shape Analysis:
This category of questions focuses on the model's ability to recognize and categorize objects based on their geometric shapes. Questions involve identifying items with specific shapes, such as cylindrical, circular  or rectangular objects, which are common in waste sorting processes.
\item Complex and Cluttered Environment:
This category includes questions to  evaluate the model's performance in recognizing and reasoning about the environment in which waste is found. Model evaluates whether waste is in an indoor or outdoor setting. It includes questions that require comprehensive image analysis.
\item Condition Evaluation:
In this category, the model must evaluate the condition of waste items. This includes assessing whether items are intact, twisted, clean or dirty. The questions are designed to test the model's ability to make nuanced judgments about the state of objects.
\item Similarity Metric:
These questions  require the model to compare and determine the similarity between different waste items. For example, identifying items that belong to the same category or have similar features. It assesses the model's ability to draw comparisons and make associations based on visual features, robustness in recognizing objects in challenging settings, and adaptability to varying conditions.
\item Combined Classification and Counting:
This category merges classification and counting tasks, requiring the model to not only classify multiple items in a scene but also provide accurate counts for each category. 
This combined approach tests the model's capability to perform multiple reasoning tasks simultaneously.
\end{itemize}
These question types  present in our dataset help to  rigorously test the capabilities of VLLMs in handling the intricacies of waste classification in complex and cluttered environments.
\subsection{Building Waste Bench Benchmark}

\begin{figure*}
  \centering
   \includegraphics[width=1\textwidth]{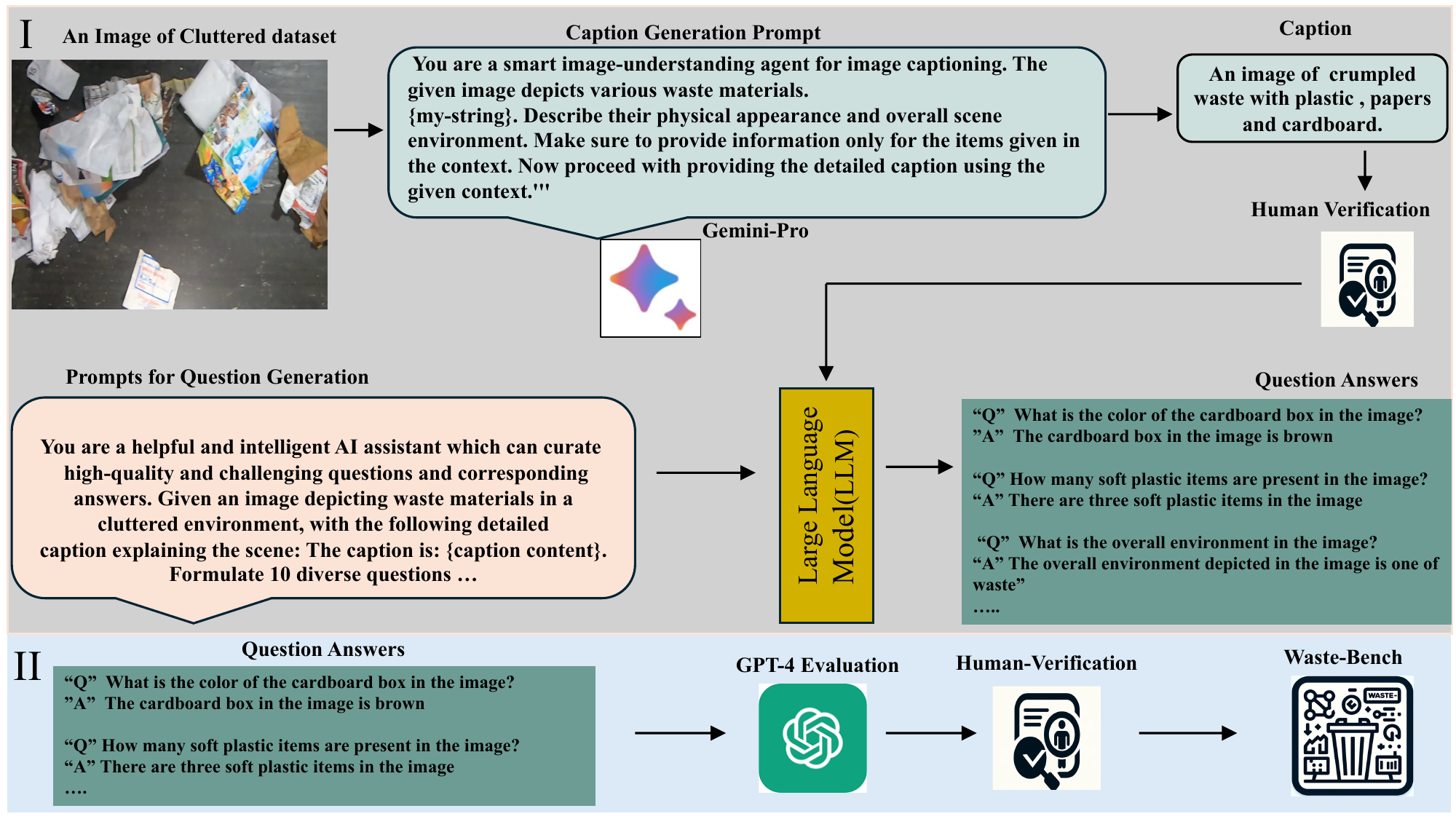}
  \caption{Step I: Gemini-Pro generates detailed waste image captions, verified by human annotators. Step II: Nearly 10k diverse questions are generated from these captions, evaluated by GPT-4, and verified by humans.}
  \label{fig:Method}
\end{figure*}
The Waste-Bench benchmark is carefully constructed through a four-step process using a dataset of 952 images. Initially, 11,424 Question/Answer (Q/A) pairs are generated, capturing information from the images. With filtering process given in Stage 1, this number is reduced to 9,520, ensuring relevance and quality. A focused refinement filtered out   1,920 Q/A pairs, representing approximately 20\% of the original set.  Each step is presented in detail below, and can be visually explored in Figure \ref{fig:Method}. 
\\
\textbf{Stage 1: Data Collection and Annotation}
We thoroughly reviewed various datasets and used  ZeroWaste \cite{Bashkirova_2022_CVPR} with waste images in cluttered environment. We  pre-processed the metadata provided with the images to ensure accurate representation of the categories assigned to each image. 
Following image collection, descriptive captions were generated with \textsc{Gemini-Pro} v1.5 (captioning) and \textsc{Gemini-Pro} v1.0 (49.45 \% precision, classification baseline).  
Two expert annotators independently reviewed each caption; only captions in which both agreed every class mention was correct were retained, otherwise they were corrected or discarded.  
Inter-rater reliability was substantial (Cohen’s $\kappa=0.78$, 95 \% CI 0.73–0.83), confirming the consistency of the process.  

\begin{itemize}
  \item \textbf{Semantic relevance.} Caption must refer only to objects actually present; any incorrect or missing class label triggered correction or rejection.%
  \item \textbf{Clarity and fluency.} Language was edited for succinct, unambiguous description.%
  \item \textbf{Technical accuracy.} Quantities, materials and spatial relations were verified against the image.%
\end{itemize}

This human-in-the-loop filtering produced concise, context-rich descriptions that remain competitive with state-of-the-art systems.

The prompt used to generate captions is provided in Figure \ref{fig:Method}. These prompts included ground-truth information (e.g., class names, categories, and masks) from the dataset's JSON annotations to guide LLMs in producing contextually accurate outputs.
  \\
\textbf{Stage 2: Generation of questions and answers}
 \begin{figure}[htb]
  \centering
  \includegraphics[width=.38\textwidth]{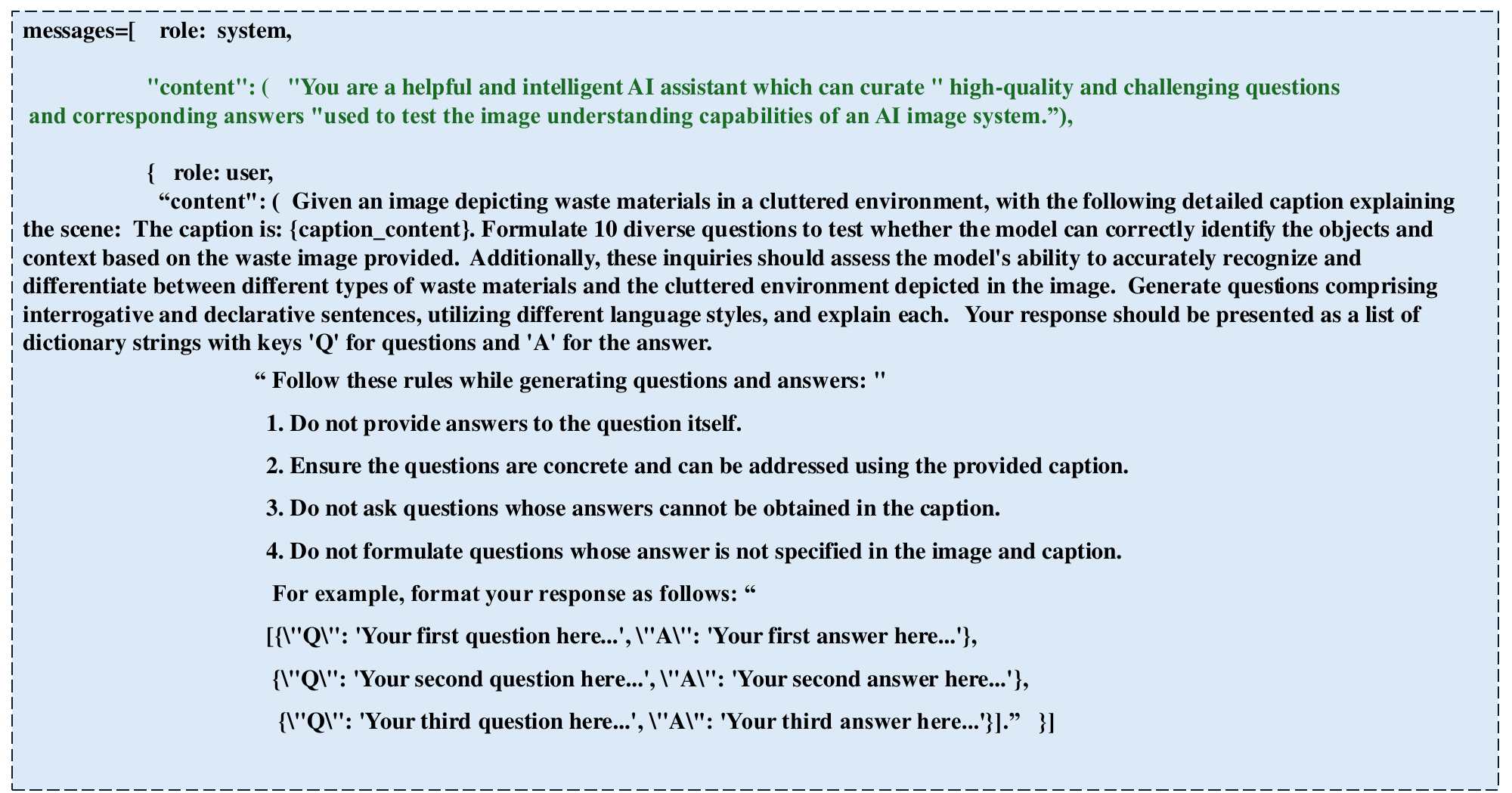}
  \caption{Prompts Used for Generating Question-Answer Pairs.}
  \label{fig:Question_Answer}
\end{figure} 
Inspired by human interaction in day-to-day life, we aim to simulate a similar style of interaction with VLLMs by curating open-ended QA pairs to evaluate these models for robustness and reasoning. We feed detailed ground-truth image captions to GPT-3.5, which are utilized to generate open-ended questions covering both reasoning and robustness aspects.
 
The questions designed go beyond basic image comprehension, requiring complex logical inference and contextual understanding. These questions test the model's ability to classify objects by recognition, color, shape, and other relevant aspects in complex settings, ensuring accurate and appropriate responses. Prompt used for curating QA pairs is mentioned in Figure \ref{fig:Question_Answer}.\\ 
\textbf{Stage 3: QA Pairs Filtration}

After generating QA pairs, a human-in-the-loop review involving two human assistants identified approximately 20\% of the pairs as noisy. These noisy pairs included irrelevant, unanswerable, or repetitive questions, such as those with answers embedded within the questions. To address these issues, an exhaustive filtering process was conducted, ensuring that the QA pairs met the relevance and alignment criteria based on the image evaluation.

For the review process, we applied similar rules as those used for caption generation. Two human assistants reviewed the question-answer pairs based on the following criteria:
\begin{itemize}
    \item QA pairs needed to be related to verified captions, both assistants agreeing that the content was relevant to the image 80\%. We now report Cohen's $\kappa = 0.78$ (95\% CI [0.73--0.83], $n = 1000$), in a random sample of 1{,}000 Q/A pairs that indicate substantial agreement between the two annotators~\cite{landis1977measurement}. The reliability of the inter-annotator on a subset of 1{,}000 items was substantial ($\kappa = 0.78$) as given in Appendix~A.1.
    \item The language was checked for clarity.
    \item The accuracy and relevance of the responses was verified.
\end{itemize}

This process ensured that only relevant, accurate, and clear question-answer pairs were retained, resulting in a curated set of 9,552 high-quality QA pairs. These pairs provide a robust foundation for the Waste-Bench benchmark. Appendix \ref{Data Filteration} provides a quantitative overview of the results.

\textbf{Stage 4: Evaluation Procedure}
\begin{figure}
  \centering
   \includegraphics[width=1\linewidth]{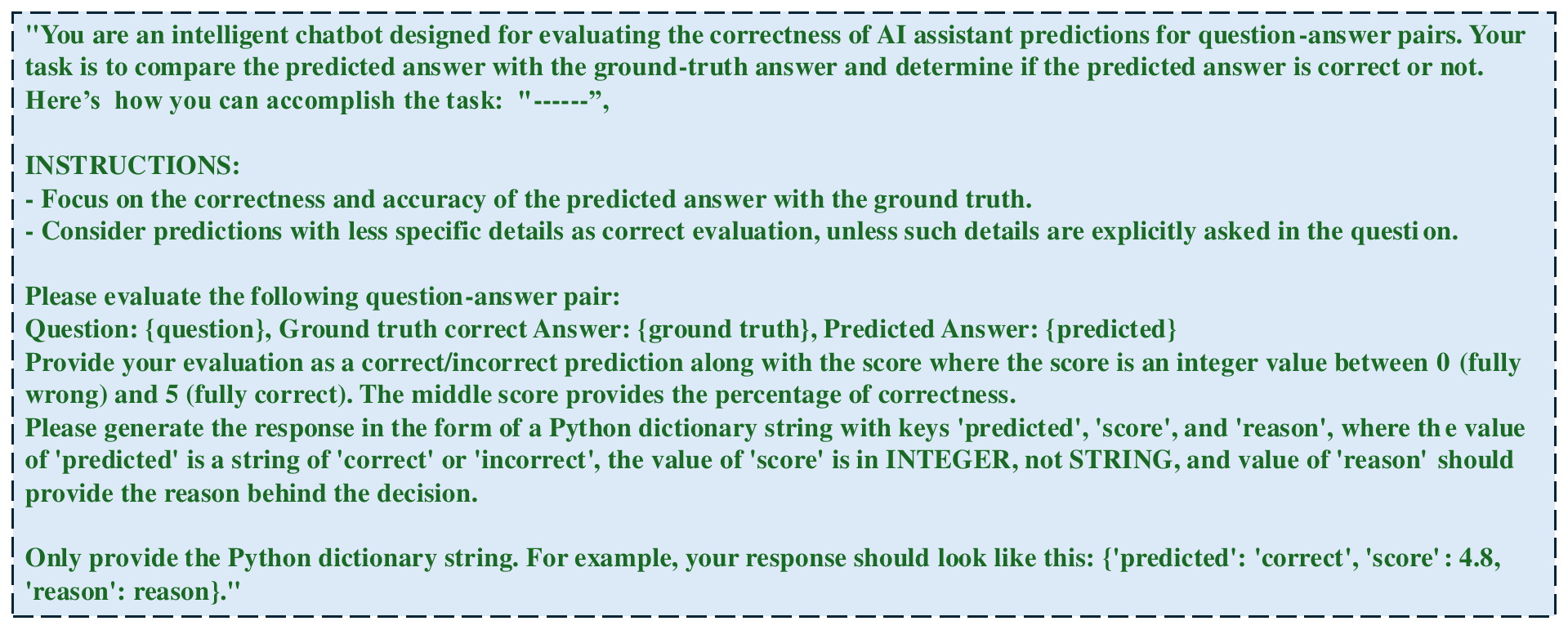}
  \caption{Evaluation prompt used.}
  \label{fig:combined_prompts}
\end{figure}
Previous methods like MM-VET\cite{yu2023mm} and SEEDBENCH \cite{li2023seed} have used LLMs as judges for open-ended QA benchmarks. We follow a similar approach, employing GPT-4 to evaluate the correctness of VLLM predictions against ground-truth answers. VLLMs generate predictions based on image-question pairs, which are then assessed by GPT-4 through binary judgments, with reasoning provided for each decision. 
The evaluation prompt as given in Figure   \ref  {fig:combined_prompts},
 used in our study was designed to guide the LLMs in assessing the accuracy and quality of the responses generated by VLLMs on the Waste-Bench dataset. This prompt provided the LLMs with specific instructions to compare the model-generated answers with ground-truth answers, make binary correctness judgments. The prompt also emphasized the importance of providing reasoning for each evaluation, ensuring that the judgments were not only accurate but also interpretable and consistent. To ensure accuracy, two assistants reviewed the evaluation results.  To validate the performance across all models, we observed a high consistency between GPT-4 and human evaluations, as given in Table \ref{tab:human-gpt}
 below.

\begin{table}[h!]
\centering

\setlength{\tabcolsep}{10pt} 
\arrayrulecolor{gray} 
\resizebox{\columnwidth}{!}{ 
\begin{tabular}{lcc|cc}
\rowcolor[gray]{0.9} 
\toprule
 & \multicolumn{2}{c|}{\textbf{GPT}} & \multicolumn{2}{c}{\textbf{Human}} \\
\midrule
\textbf{Model} & \textbf{CogVLM} & \textbf{InstructBLIP} & \textbf{InstructBLIP} & \textbf{CogVLM} \\
\textbf{Performance} & 45\% & 59\% & 63\% & 46\% \\
\bottomrule
\end{tabular}
}
\caption{Comparison of model performance between GPT and Human evaluations across different models.}
\label{tab:human-gpt}
\end{table}



\section{Performance Evaluation on Waste-Bench}
\begin{table*}[ht]
\centering
\begin{tabular}{l l l c}
\toprule
\rowcolor[gray]{0.9}
\textbf{Model}        & \textbf{Version}           & \textbf{LLM}               & \textbf{Accuracy (\%)} \\
\midrule
GPT-4                 & GPT-4o                     & Proprietary LLM            & 57.52                  \\
Gemini                & Gemini-1.0 Pro             & Proprietary LLM            & 49.45                  \\
InstructBLIP          & BLIP-2\_Vicuna\_Instruct   & Vicuna-7B                  & 48.58                  \\
LLaVA                 & LLaVA-1.6                  & Vicuna-7B                  & 47.45                  \\
Qwen-VL  & Qwen-VL-Chat               & Qwen-7B         & 41.30                   \\
CogVLM                & CogVLM-chat-v1.1           & Vicuna-7B                  & 41.58                  \\
MiniGPT-4             & MiniGPT-4                  & Vicuna-7B                  & 36.40                  \\

\midrule
\textbf{Human Upper Bound} & N/A                   & N/A                        & \textbf{81.20}         \\
\bottomrule
\end{tabular}
\caption{Evaluation results VLLMs  highlighting open-source and closed-source models.}
\label{tab:tab-results1}
\end{table*}

\begin{table*}[htbp]
\centering
\resizebox{\textwidth}{!}{%
\begin{tabular}{lcccccccc}
\rowcolor[gray]{0.9}
\hline
\textbf{Question Category}                & \textbf{GPT-4} & \textbf{Gemini} & \textbf{InstructBLIP} & \textbf{LLAVA} & \textbf{Qwen-VL} & \textbf{CogVLM} & \textbf{MiniGPT-4} 
\\ \hline
Single Class Classification              & 49.00          & 38.00           & 46.00                 & 35.00          & 28.50            & 36.50            & 22.00             
\\ 
Multiclass Categorization                & 54.00          & 44.00           & 36.50                 & 37.00          & 34.00            & 30.50            & 32.00              
\\ 
Counting                                 & 60.00          & 52.00           & 50.00                 & 45.50          & 43.00            & 40.50            & 31.00              
\\ 
Color Diversity                          & 42.00          & 35.00           & 39.00                 & 48.00          & 38.00            & 27.50            & 30.00              
\\ 
Geometric Shape Analysis                 & 55.00          & 49.00           & 44.00                 & 41.50          & 45.50            & 39.00            & 36.50              
\\ 
Complex and Cluttered Environment        & 38.00          & 42.00           & 52.00                 & 58.00          & 51.00            & 47.00            & 39.00              
\\ 
Condition Evaluation                     & 60.00          & 57.00           & 48.50                 & 49.50          & 38.00            & 33.00            & 35.00              
\\ 
Similarity Metric                        & 53.50          & 47.00           & 38.50                 & 56.00          & 44.50            & 50.50            & 29.00             
\\ 
Combined Classification and Counting     & 44.00          & 48.00           & 53.00                 & 44.50          & 39.00            & 41.00            & 36.00              
\\ \hline
\end{tabular}%
}
\caption{Comparison of different models across question categories using weighted average scores, highlighting the relative performance of open-source and closed-source models.}
\label{tab:model_comparison}
\end{table*}

Both open-source and closed-source models were explored and selected for evaluation. In total, seven models were evaluated. Among the open-source models, five recent VLLMs were included: InstructBLIP, LLaVA-1.6, CogVLM, Qwen-VL, and MiniGPT-4. For closed-source models, GPT-4o and Gemini-Pro were used. Our work focuses on evaluating existing VLLMs to highlight their limitations in cluttered environments. While VLLMs are costly to train, our evaluation reveals key challenges, and future work will address issues like hallucination and robustness for better performance in complex tasks.

\subsection{Main Experiments on Waste-Bench}
All models were used in their pre-trained state to ensure a fair comparison across different architectures, detail given in in Appendix Table \ref{tab:settings}. Given the diversity of the models employed, specific hyperparameter tuning was not performed for individual models; instead, the focus was on evaluating their inherent capabilities. Each model was assessed under consistent conditions, using a single NVIDIA 24GB GPU to run the experiments, ensuring uniformity in computational resources across the tasks. 
\begin{figure*}
  \centering
   \includegraphics[width=.9\textwidth]{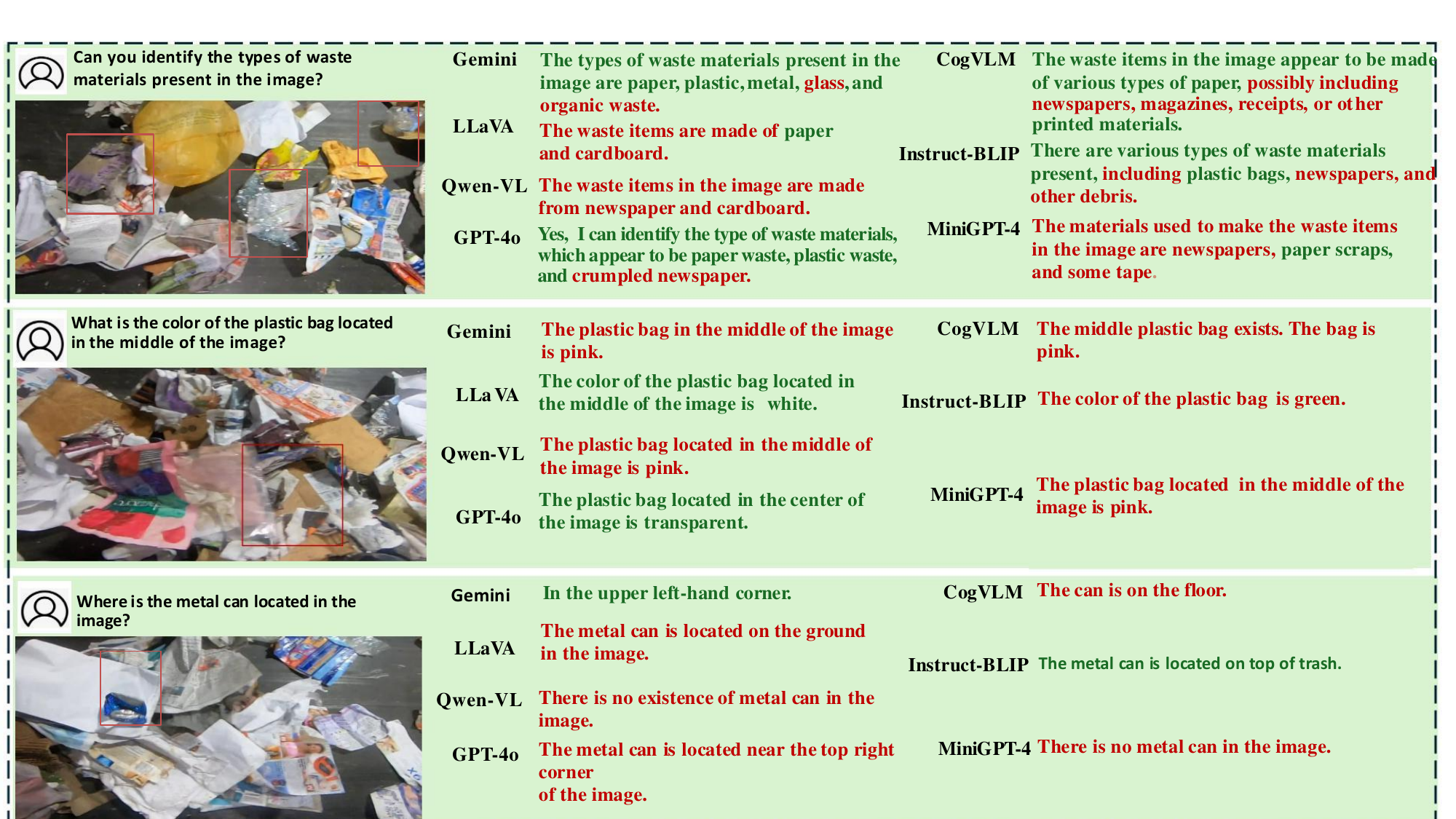}
   \caption{Qualitative results illustrating models struggling with identifying shapes, colors, and recognizing rare classes within cluttered scenes, indicating areas for further investigation and improvement.} 
  \label{fig:evaluation}
\end{figure*}
In Table  \ref{tab:tab-results1}, we present the evaluation results of diverse range of models including five open source, two closed source and human upper bound to provide comprehensive benchmark . All evaluations were conducted according to the settings specified officially as discussed in Appendix  \ref{Evaluation}
and Table \ref{tab:settings}.  
VLLMs find it challenging to perform well and thus show inferior performance  when evaluated on the Waste-Bench dataset, particularly in cluttered scenes with deformed shaped objects. Interestingly, the performance of models like  LLaVA-1.6, and InstructBLIP is relatively higher compared to models such as Qwen-VL and MiniGPT-4. For instance, Gemini achieves an accuracy of 49.45\% , however MiniGPT-4 suffers severely with these particularly challenging conditions and thus under perform. 
The Gpt-4o model surpasses the performance of all models and achieves high gains compared to other models. However, it still remains at the lower end of performance for this type of dataset, with an accuracy around 57\%. GPT-4o handles cluttered scenes with deformed shaped objects, better than others, indicating a more sophisticated understanding of complex visual contents. 
The Table \ref{tab:model_comparison} compares the performance of various VLLMs across different waste classification tasks. GPT-4 performs well in most categories, especially in Counting (60.00) and Condition Evaluation (60.00), while MiniGPT-4 shows weaker results, particularly in Single Class Classification (22.00). Models like Gemini and LLAVA exhibit moderate performance, with LLAVA excelling in Condition Evaluation (58.00). The values are rounded to whole numbers for simplicity and clarity.

Using the accuracies in Table 3, we compute the number of errors for each question category \(c\) as
\[
\mathrm{errors}_c \;=\; 952 \,\Bigl(1-\tfrac{\mathrm{accuracy}_c}{100}\Bigr),
\]
where \(952\) is the number of questions in that category.  
Summing the resulting counts over the seven evaluated VLMs gives:  
colour mis-identification \(= 4\,194\) (12.2\%),  
single-class slips \(= 4\,236\) (12.3\%),  
multiclass confusion \(= 4\,113\) (12.0\%),  
complex-scene reasoning \(= 3\,551\) (10.3\%),  
geometric-shape confusion \(= 3\,708\) (10.8\%),  
counting mismatches \(= 3\,599\) (10.5\%),  
condition mis-classification \(= 3\,608\) (10.5\%),  
similarity errors \(= 3\,627\) (10.5\%),  
and combined class\,+\,count errors \(= 3\,756\) (10.9\%),  
for a grand total of \(34\,391\) errors. 

\section{Key Highlights and Qualitative Results}
The evaluation of VLLMs on the Waste-Bench benchmark reveals critical insights valuable for future model development, focusing on model performance under various conditions and highlighting strengths and areas for improvement.\\
\textbf{Real-World Waste Classification Challenges:\\
} Models that perform well on simplified environments often struggle with the complexities of Waste-Bench, particularly when it comes to counting irregularly shaped objects or accurately identifying colors in cluttered scenes. For instance, as illustrated in Figure \ref{fig:evaluation}, Q2, a model incorrectly predicted the color of a plastic bag due to a colored paper beneath it, highlighting  challenges of real-world waste classification, where objects are frequently stacked or partially obscured to make it difficult to predict. Models often struggle with correctly identifying colors in cluttered scenes due to the lack of real-world complexity in their training data. Enhancing training with diverse and realistic samples could help improve their accuracy and robustness in complex environment.\\
\textbf{Challenges in Rare Class Recognition:}
Models often struggle to accurately recognize and classify less frequent categories in cluttered scenes, particularly when objects are deformed. As seen in Q3, models mislocate or miss the metal, highlighting the need for improved training on diverse variety of deformed object shapes in cluttered environment which are often encountered in real world streams.\\
\textbf{Weak Classification in Cluttered Environments:
}The responses in Question 1 highlight key challenges in accurate material identification, particularly in scenes where objects are partially obscured. For example, while some models like GPT-4o correctly identify a range of materials, others like LLaVA and Qwen-VL struggled, with differentiating between visually similar objects, leading to incomplete or incorrect classifications. This inconsistency underscores the need for further refinement of VLLMs to improve their robustness in real-world applications, such as automated waste management, where precise identification is critical. Further insights are given in Appendix \ref{Insight}. \\ \textbf{Potential Data Leakage:}
This is dataset which is maintained by independent research group and cannot be obtained by using web crawling techniques which VLLMS use to curate their datasets.

\section{Validation and Comparison Across Other BenchMarks }
The Table \ref{tab:benchmark-comparison-final} compares the accuracy of various VLLMs across various benchmarks. Notably, the table illustrates the diverse challenges posed by each benchmark, with Waste-Bench offering a unique set of difficulties due to its focus on cluttered scenes with deformed objects. The performance of models such as LLaVA, InstructBLIP, and Qwen-VL shows a noticeable drop in accuracy on Waste-Bench compared to SEED-Bench and MV-Bench. 
This highlights the increased complexity and difficulty in real-world waste classification scenarios and need to optimize current models for the unique challenges of waste classification.

\section{Conclusion}
In this paper, we evaluated various VLLMs in complex environments with deformed objects, revealing significant weaknesses in the identification of shapes, colors, and locations. We introduced the Waste-Bench benchmark, which features multiple categories to enable a comprehensive validation of these models. The Waste-Bench benchmark provides a robust framework for assessing VLLMs in challenging conditions, aiding in the development of more resilient and accurate models for real-world applications like waste segregation and autonomous waste management.

\begin{table}[!htbp] 
\centering
\tiny
\begin{tabular}{lcccc}
\rowcolor[gray]{0.7}
\toprule
\textbf{Model} & \textbf{MM-VET} & \textbf{MV-Bench} & \textbf{SEED-Bench} & \textbf{Waste-Bench} \\
\midrule
GPT-4          & -               & -                 & -                   & 57.5 \\
Gemini         & -               & -                 & -                   & 49.4 \\
InstructBLIP   & 69.9            & 51.0              & 61.7                & 48.6 \\
LLaVA          & 46.6            & 53.0              & 66.7                & 47.4 \\
Qwen-VL        & -               & 73.0              & 54.8                & 41.3 \\
CogVLM         & -               & -                 & -                   & 41.6 \\
MiniGPT-4      & 47.9            & 29.5              & 49.2                & 36.4 \\
Human Upper Bound & -            & -                 & -                   & 81.2 \\
\bottomrule
\end{tabular}

\caption{Comparison of VLLM recognition performance across different benchmarks in terms of accuracy.}
\label{tab:benchmark-comparison-final}

\noindent {Note:} In this table, -- values indicate results not provided.
\end{table}

\textbf{Limitations}
Our study, though comprehensive, has some limitations.
 The scope of our evaluation was limited to a specific set of cluttered environments, which may not fully represent the variety of real-world scenarios. In addition, the models were tested under controlled conditions and their performance in more dynamic and unpredictable settings remains to be explored. We tested models on a variety of questions to ensure robust testing for our evaluation purposes, accuracy and score were calculated and  seemed sufficient, showcasing the robustness of our approach. Incorporating additional evaluation methods in future work could provide a more complete understanding. Despite these limitations, our findings offer valuable insight and a strong foundation to advance research in this area.\\
 \textbf{Ethics Statement}
We constructed this dataset based on images given in the zwaste-f dataset \cite{Bashkirova_2022_CVPR}. We constructed this data set based on images provided in the Zerowaste-F dataset (Bashkirova et al., 2022). This data set includes various images of waste in cluttered environments to simulate real-world conditions. Some images contain identifiable objects, but we ensured that no personal identification details are included. When used properly, our image and annotation dataset provides significant value for evaluating waste classification models.

\bibliography{custom}

\begin{thebibliography}{34}
\providecommand{\natexlab}[1]{#1}

\bibitem[{Achiam et~al.(2023)Achiam, Adler, Agarwal, Ahmad, Akkaya, Aleman, Almeida, Altenschmidt, Altman, Anadkat et~al.}]{achiam2023gpt}
Josh Achiam, Steven Adler, Sandhini Agarwal, Lama Ahmad, Ilge Akkaya, Florencia~Leoni Aleman, Diogo Almeida, Janko Altenschmidt, Sam Altman, Shyamal Anadkat, et~al. 2023.
\newblock Gpt-4 technical report.
\newblock \emph{arXiv preprint arXiv:2303.08774}.

\bibitem[{Alayrac et~al.(2022)Alayrac, Donahue, Luc, Miech, Barr, Hasson, Lenc, Mensch, Millican, Reynolds et~al.}]{alayrac2022flamingo}
Jean-Baptiste Alayrac, Jeff Donahue, Pauline Luc, Antoine Miech, Iain Barr, Yana Hasson, Karel Lenc, Arthur Mensch, Katherine Millican, Malcolm Reynolds, et~al. 2022.
\newblock Flamingo: a visual language model for few-shot learning.
\newblock \emph{Advances in neural information processing systems}, 35:23716--23736.

\bibitem[{Anil et~al.(2023)Anil, Dai, Firat, Johnson, Lepikhin, Passos, Shakeri, Taropa, Bailey, Chen et~al.}]{anil2023palm}
Rohan Anil, Andrew~M Dai, Orhan Firat, Melvin Johnson, Dmitry Lepikhin, Alexandre Passos, Siamak Shakeri, Emanuel Taropa, Paige Bailey, Zhifeng Chen, et~al. 2023.
\newblock Palm 2 technical report.
\newblock \emph{arXiv preprint arXiv:2305.10403}.

\bibitem[{Bai et~al.(2023)Bai, Bai, Chu, Cui, Dang, Deng, Fan, Ge, Han, Huang, Hui, Ji, Li, Lin, Lin, Liu, Liu, Lu, Lu, Ma, Men, Ren, Ren, Tan, Tan, Tu, Wang, Wang, Wang, Wu, Xu, Xu, Yang, Yang, Yang, Yang, Yao, Yu, Yuan, Yuan, Zhang, Zhang, Zhang, Zhang, Zhou, Zhou, Zhou, and Zhu}]{qwen}
Jinze Bai, Shuai Bai, Yunfei Chu, Zeyu Cui, Kai Dang, Xiaodong Deng, Yang Fan, Wenbin Ge, Yu~Han, Fei Huang, Binyuan Hui, Luo Ji, Mei Li, Junyang Lin, Runji Lin, Dayiheng Liu, Gao Liu, Chengqiang Lu, Keming Lu, Jianxin Ma, Rui Men, Xingzhang Ren, Xuancheng Ren, Chuanqi Tan, Sinan Tan, Jianhong Tu, Peng Wang, Shijie Wang, Wei Wang, Shengguang Wu, Benfeng Xu, Jin Xu, An~Yang, Hao Yang, Jian Yang, Shusheng Yang, Yang Yao, Bowen Yu, Hongyi Yuan, Zheng Yuan, Jianwei Zhang, Xingxuan Zhang, Yichang Zhang, Zhenru Zhang, Chang Zhou, Jingren Zhou, Xiaohuan Zhou, and Tianhang Zhu. 2023.
\newblock Qwen technical report.
\newblock \emph{arXiv preprint arXiv:2309.16609}.

\bibitem[{Bashkirova et~al.(2022)Bashkirova, Abdelfattah, Zhu, Akl, Alladkani, Hu, Ablavsky, Calli, Bargal, and Saenko}]{Bashkirova_2022_CVPR}
Dina Bashkirova, Mohamed Abdelfattah, Ziliang Zhu, James Akl, Fadi Alladkani, Ping Hu, Vitaly Ablavsky, Berk Calli, Sarah~Adel Bargal, and Kate Saenko. 2022.
\newblock Zerowaste dataset: Towards deformable object segmentation in cluttered scenes.
\newblock In \emph{Proceedings of the IEEE/CVF Conference on Computer Vision and Pattern Recognition (CVPR)}, pages 21147--21157.

\bibitem[{Chou et~al.(2020)Chou, Chao, Lai, Sun, and Yang}]{9093452}
Shih-Han Chou, Wei-Lun Chao, Wei-Sheng Lai, Min Sun, and Ming-Hsuan Yang. 2020.
\newblock \href {https://doi.org/10.1109/WACV45572.2020.9093452} {Visual question answering on 360° images}.
\newblock In \emph{2020 IEEE Winter Conference on Applications of Computer Vision (WACV)}, pages 1596--1605.

\bibitem[{Chung et~al.(2024)Chung, Hou, Longpre, Zoph, Tay, Fedus, Li, Wang, Dehghani, Brahma et~al.}]{chung2024scaling}
Hyung~Won Chung, Le~Hou, Shayne Longpre, Barret Zoph, Yi~Tay, William Fedus, Yunxuan Li, Xuezhi Wang, Mostafa Dehghani, Siddhartha Brahma, et~al. 2024.
\newblock Scaling instruction-finetuned language models.
\newblock \emph{Journal of Machine Learning Research}, 25(70):1--53.

\bibitem[{Dai et~al.(2024)Dai, Li, Li, Tiong, Zhao, Wang, Li, Fung, and Hoi}]{dai2024instructblip}
Wenliang Dai, Junnan Li, Dongxu Li, Anthony Meng~Huat Tiong, Junqi Zhao, Weisheng Wang, Boyang Li, Pascale~N Fung, and Steven Hoi. 2024.
\newblock Instructblip: Towards general-purpose vision-language models with instruction tuning.
\newblock \emph{Advances in Neural Information Processing Systems}, 36.

\bibitem[{Fang et~al.(2023)Fang, Sun, Wang, Huang, Wang, and Cao}]{fang2023eva}
Yuxin Fang, Quan Sun, Xinggang Wang, Tiejun Huang, Xinlong Wang, and Yue Cao. 2023.
\newblock Eva-02: A visual representation for neon genesis.
\newblock \emph{arXiv preprint arXiv:2303.11331}.

\bibitem[{Feng et~al.(2022)Feng, Yang, Chen, Chen, and Li}]{feng2022intelligent}
Zhicheng Feng, Jie Yang, Lifang Chen, Zhichao Chen, and Linhong Li. 2022.
\newblock An intelligent waste-sorting and recycling device based on improved efficientnet.
\newblock \emph{International Journal of Environmental Research and Public Health}, 19(23):15987.

\bibitem[{Ganin and Lempitsky(2016)}]{ganin2016dann}
Yaroslav Ganin and Victor Lempitsky. 2016.
\newblock Unsupervised domain adaptation by backpropagation.
\newblock \emph{Journal of Machine Learning Research}, 17(59):1--35.

\bibitem[{Havrilla et~al.(2024)Havrilla, Dai, O'Mahony, Oostermeijer, Zisler, Albalak, Milo, Raparthy, Gandhi, Abbasi et~al.}]{havrilla2024surveying}
Alex Havrilla, Andrew Dai, Laura O'Mahony, Koen Oostermeijer, Vera Zisler, Alon Albalak, Fabrizio Milo, Sharath~Chandra Raparthy, Kanishk Gandhi, Baber Abbasi, et~al. 2024.
\newblock Surveying the effects of quality, diversity, and complexity in synthetic data from large language models.
\newblock \emph{arXiv preprint arXiv:2412.02980}.

\bibitem[{Hu et~al.(2023)Hu, Yuan, Wen, Lu, and Li}]{hu2023rsgpt}
Yuan Hu, Jianlong Yuan, Congcong Wen, Xiaonan Lu, and Xiang Li. 2023.
\newblock Rsgpt: A remote sensing vision language model and benchmark.
\newblock \emph{arXiv preprint arXiv:2307.15266}.

\bibitem[{Landis and Koch(1977)}]{landis1977measurement}
J.~Richard Landis and Gary~G. Koch. 1977.
\newblock \href {https://doi.org/10.2307/2529310} {The measurement of observer agreement for categorical data}.
\newblock \emph{Biometrics}, 33(1):159--174.

\bibitem[{Li et~al.(2023)Li, Wang, Wang, Ge, Ge, and Shan}]{li2023seed}
Bohao Li, Rui Wang, Guangzhi Wang, Yuying Ge, Yixiao Ge, and Ying Shan. 2023.
\newblock Seed-bench: Benchmarking multimodal llms with generative comprehension.
\newblock \emph{arXiv preprint arXiv:2307.16125}.

\bibitem[{Li et~al.(2024)Li, Wang, He, Li, Wang, Liu, Wang, Xu, Chen, Luo et~al.}]{li2024mvbench}
Kunchang Li, Yali Wang, Yinan He, Yizhuo Li, Yi~Wang, Yi~Liu, Zun Wang, Jilan Xu, Guo Chen, Ping Luo, et~al. 2024.
\newblock Mvbench: A comprehensive multi-modal video understanding benchmark.
\newblock In \emph{Proceedings of the IEEE/CVF Conference on Computer Vision and Pattern Recognition}, pages 22195--22206.

\bibitem[{Liu et~al.(2023)Liu, Li, Li, and Lee}]{liu2023improvedllava}
Haotian Liu, Chunyuan Li, Yuheng Li, and Yong~Jae Lee. 2023.
\newblock Improved baselines with visual instruction tuning.

\bibitem[{Liu et~al.(2024)Liu, Li, Li, Li, Zhang, Shen, and Lee}]{liu2024llavanext}
Haotian Liu, Chunyuan Li, Yuheng Li, Bo~Li, Yuanhan Zhang, Sheng Shen, and Yong~Jae Lee. 2024.
\newblock \href {https://llava-vl.github.io/blog/2024-01-30-llava-next/} {Llava-next: Improved reasoning, ocr, and world knowledge}.

\bibitem[{Madry et~al.(2018)Madry, Makelov, Schmidt, Tsipras, and Vladu}]{madry2018pgd}
Aleksander Madry, Aleksandar Makelov, Ludwig Schmidt, Dimitris Tsipras, and Adrian Vladu. 2018.
\newblock Towards deep learning models resistant to adversarial attacks.
\newblock In \emph{Proceedings of the International Conference on Learning Representations}.

\bibitem[{Mao et~al.(2021)Mao, Chen, Wang, and Lin}]{mao2021recycling}
Wei-Lung Mao, Wei-Chun Chen, Chien-Tsung Wang, and Yu-Hao Lin. 2021.
\newblock Recycling waste classification using optimized convolutional neural network.
\newblock \emph{Resources, Conservation and Recycling}, 164:105132.

\bibitem[{Meng et~al.(2022)Meng, Jiang, Wang, and Wang}]{meng2022mobilenet}
Jing Meng, Ping Jiang, Jianmin Wang, and Kai Wang. 2022.
\newblock A mobilenet-ssd model with fpn for waste detection.
\newblock \emph{Journal of Electrical Engineering \& Technology}, 17(2):1425--1431.

\bibitem[{OpenAI(2024)}]{openai2024gpt4o}
OpenAI. 2024.
\newblock \href {https://openai.com/index/hello-gpt-4o/} {Hello gpt-4o}.
\newblock Accessed: 2024-05-26.

\bibitem[{Penedo et~al.(2023)Penedo, Malartic, Hesslow, Cojocaru, Cappelli, Alobeidli, Pannier, Almazrouei, and Launay}]{refinedweb}
Guilherme Penedo, Quentin Malartic, Daniel Hesslow, Ruxandra Cojocaru, Alessandro Cappelli, Hamza Alobeidli, Baptiste Pannier, Ebtesam Almazrouei, and Julien Launay. 2023.
\newblock \href {https://arxiv.org/abs/2306.01116} {The {R}efined{W}eb dataset for {F}alcon {LLM}: outperforming curated corpora with web data, and web data only}.
\newblock \emph{arXiv preprint arXiv:2306.01116}.

\bibitem[{Redmon(2016)}]{redmon2016you}
J~Redmon. 2016.
\newblock You only look once: Unified, real-time object detection.
\newblock In \emph{Proceedings of the IEEE conference on computer vision and pattern recognition}.

\bibitem[{Reid et~al.(2024)Reid, Savinov, Teplyashin, Lepikhin, Lillicrap, Alayrac, Soricut, Lazaridou, Firat, Schrittwieser et~al.}]{reid2024gemini}
Machel Reid, Nikolay Savinov, Denis Teplyashin, Dmitry Lepikhin, Timothy Lillicrap, Jean-baptiste Alayrac, Radu Soricut, Angeliki Lazaridou, Orhan Firat, Julian Schrittwieser, et~al. 2024.
\newblock Gemini 1.5: Unlocking multimodal understanding across millions of tokens of context.
\newblock \emph{arXiv preprint arXiv:2403.05530}.

\bibitem[{Ren(2015)}]{ren2015faster}
Shaoqing Ren. 2015.
\newblock Faster r-cnn: Towards real-time object detection with region proposal networks.
\newblock \emph{arXiv preprint arXiv:1506.01497}.

\bibitem[{Shao et~al.(2023)Shao, Hu, Gao, Lei, Zhang, Meng, Xu, Huang, Li, Qiao et~al.}]{shao2023tiny}
Wenqi Shao, Yutao Hu, Peng Gao, Meng Lei, Kaipeng Zhang, Fanqing Meng, Peng Xu, Siyuan Huang, Hongsheng Li, Yu~Qiao, et~al. 2023.
\newblock Tiny lvlm-ehub: Early multimodal experiments with bard.
\newblock \emph{arXiv preprint arXiv:2308.03729}.

\bibitem[{Sun et~al.(2019)Sun, Wang, Zou, and Hua}]{sun2019testtime}
Qian Sun, Chen Wang, Yexun Zou, and Gang Hua. 2019.
\newblock \href {https://doi.org/10.1109/CVPR.2019.01125} {Test-time training for generalization to distribution shifts}.
\newblock In \emph{Proceedings of the IEEE/CVF Conference on Computer Vision and Pattern Recognition}, pages 10972--10981.

\bibitem[{Touvron et~al.(2023)Touvron, Lavril, Izacard, Martinet, Lachaux, Lacroix, Rozi{\`e}re, Goyal, Hambro, Azhar et~al.}]{touvron2023llama}
Hugo Touvron, Thibaut Lavril, Gautier Izacard, Xavier Martinet, Marie-Anne Lachaux, Timoth{\'e}e Lacroix, Baptiste Rozi{\`e}re, Naman Goyal, Eric Hambro, Faisal Azhar, et~al. 2023.
\newblock Llama: Open and efficient foundation language models.
\newblock \emph{arXiv preprint arXiv:2302.13971}.

\bibitem[{Vinyals et~al.(2015)Vinyals, Toshev, Bengio, and Erhan}]{vinyals2015show}
Oriol Vinyals, Alexander Toshev, Samy Bengio, and Dumitru Erhan. 2015.
\newblock Show and tell: A neural image caption generator.
\newblock In \emph{Proceedings of the IEEE conference on computer vision and pattern recognition}, pages 3156--3164.

\bibitem[{Xia et~al.(2024)Xia, Zhou, Yu, Hu, Zhang, Hu, and He}]{xia2024yolo}
Zhongyi Xia, Houkui Zhou, Huimin Yu, Haoji Hu, Guangqun Zhang, Junguo Hu, and Tao He. 2024.
\newblock Yolo-mtg: a lightweight yolo model for multi-target garbage detection.
\newblock \emph{Signal, Image and Video Processing}, pages 1--16.

\bibitem[{Yu et~al.(2023)Yu, Yang, Li, Wang, Lin, Liu, Wang, and Wang}]{yu2023mm}
Weihao Yu, Zhengyuan Yang, Linjie Li, Jianfeng Wang, Kevin Lin, Zicheng Liu, Xinchao Wang, and Lijuan Wang. 2023.
\newblock Mm-vet: Evaluating large multimodal models for integrated capabilities.
\newblock \emph{arXiv preprint arXiv:2308.02490}.

\bibitem[{Zheng et~al.(2023)Zheng, Chiang, Sheng, Zhou, Wu, Zhuang, Lin, Li, Li, Xing, Zhang, Gonzalez, and Stoica}]{zheng2023judging}
Lianmin Zheng, Wei-Lin Chiang, Ying Sheng, Siyuan Zhou, Zhanghao Wu, Yonghao Zhuang, Zi~Lin, Zhuohan Li, Dacheng Li, Eric~P. Xing, Hao Zhang, Joseph~E. Gonzalez, and Ion Stoica. 2023.
\newblock Judging llm-as-a-judge with mt-bench and chatbot arena.
\newblock \emph{arXiv preprint arXiv:2306.05685}.

\bibitem[{Zhu et~al.(2024)Zhu, Chen, Shen, Li, and Elhoseiny}]{zhu2023minigpt}
Deyao Zhu, Jun Chen, Xiaoqian Shen, Xiang Li, and Mohamed Elhoseiny. 2024.
\newblock \href {https://openreview.net/forum?id=1tZbq88f27} {Mini{GPT}-4: Enhancing vision-language understanding with advanced large language models}.
\newblock In \emph{The Twelfth International Conference on Learning Representations}.

\end{thebibliography}

 \appendix

\section{Appendix}
 \label{sec:appendix}

\subsection{Data Filtration}
\begin{figure*}
  \centering
   \includegraphics[width=1\textwidth]{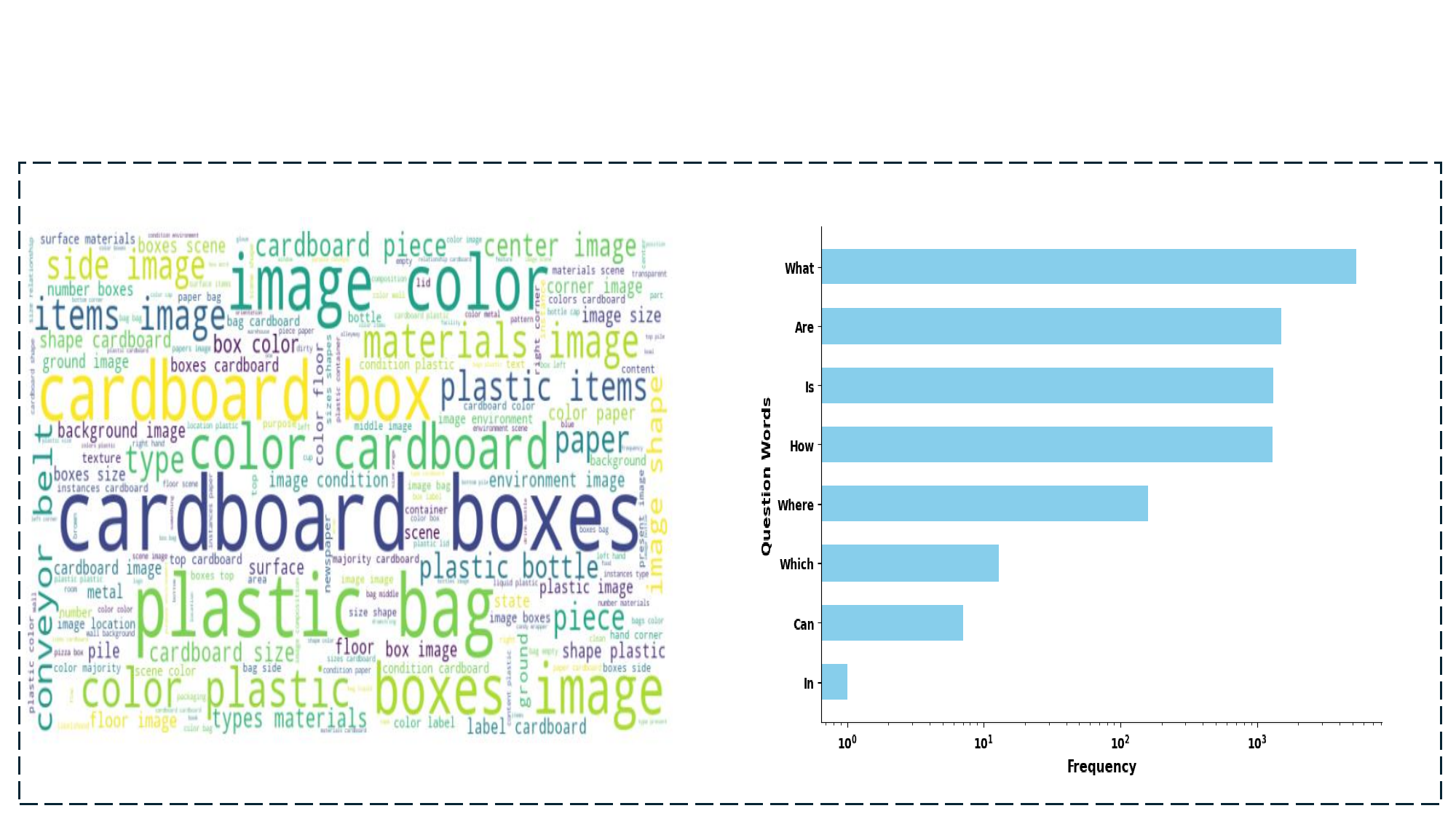}
  \caption{Waste-Bench Overview. Left: Most frequent keywords in the answer set, Right: Frequency distribution of question types. }
  \label{fig:evaluation23}
  \end{figure*}
 \label{Data Filteration}
Table \ref{tab:filter} presents an overview of the dataset statistics, including the total number of images and question-answer (Q/A) pairs. The dataset initially contains 952 images and 11,424 Q/A pairs. However, approximately 20\% of the Q/A pairs (1,904 pairs) were filtered out, leaving a total of 9,520 updated Q/A pairs for further analysis. This filtration process ensures that the data used for evaluation is of higher quality and relevance to the task at hand
\begin{table}[htb]
\centering
\begin{tabular}{lcccc}
\toprule
\textbf{} & \textbf{Images} & \textbf{Q/A} &  \textbf{Filtered} &\textbf{Updated} \\
\midrule
 & 952 & 11424   & \textasciitilde20\% [1904] & 9520\\
\bottomrule
\end{tabular}
\caption{Dataset Statistics: Overview of Total and Filtered Question-Answer Pairs}
\label{tab:filter}
\end{table}
 \begin{figure}[!ht]
  \centering
  \includegraphics[width=.52\textwidth]{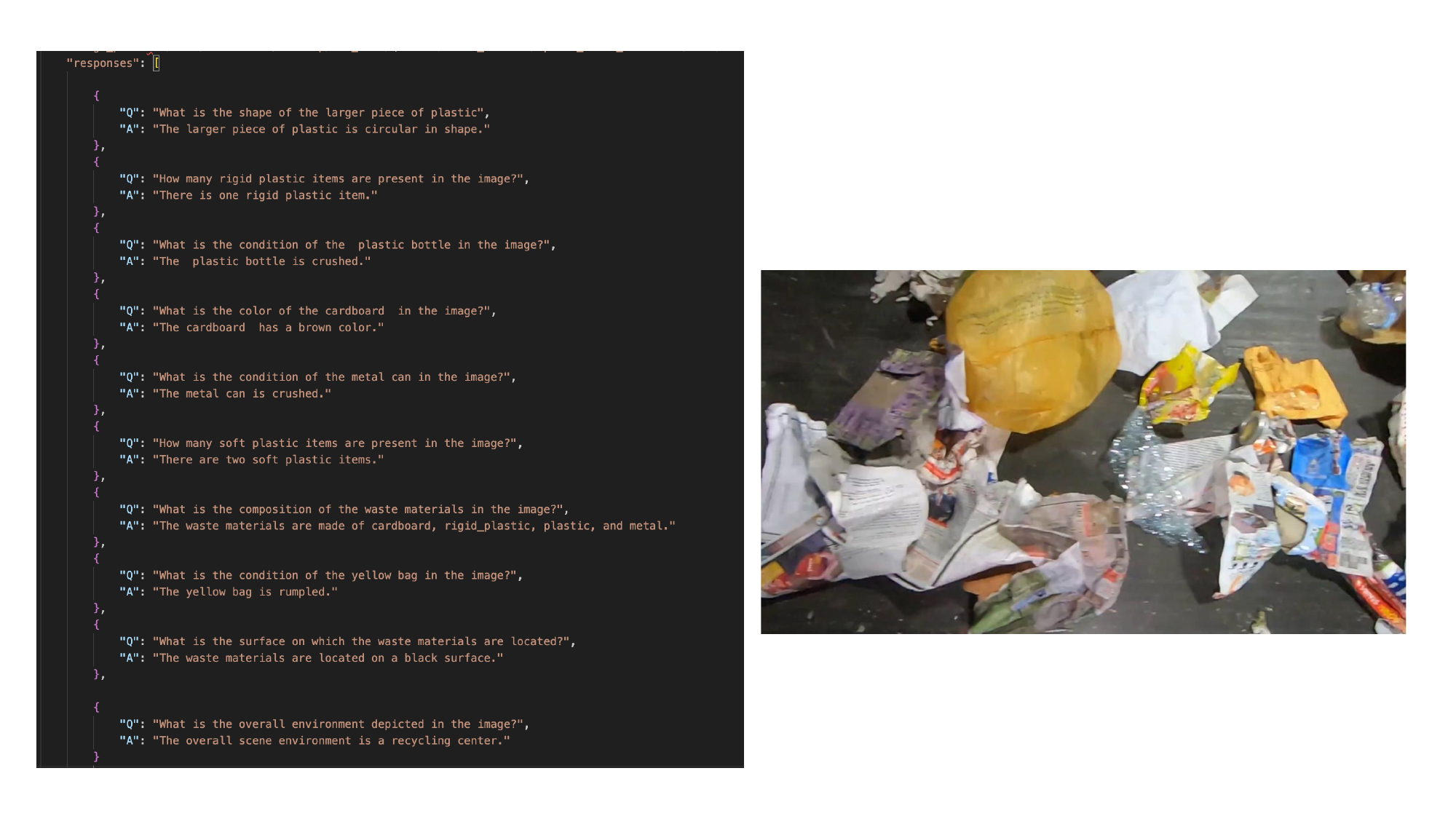}
  \caption{Q/A generation from Caption}
  \label{fig:caption}
\end{figure}





Two domain experts independently labelled a simple random sample of 1{,}000 question--answer pairs (seed = 42) with three nominal categories---Correct, Minor-Error, and Major-Error---and inter-rater reliability was assessed with Cohen's $\kappa$, yielding $\kappa = 0.78$ (95\% CI [0.73--0.83]). The statistic was computed with \texttt{sklearn}, and the confidence interval was obtained from 1{,}000 stratified bootstrap resamples. Following the Landis \& Koch (1977) interpretation, the entire interval lies in the ``substantial agreement'' band ($\kappa > 0.60$).

\subsection{WasteBench Insights}
\label{frequency}
Figure \ref{Data Filteration} provides two visualizations related to the answers in the study. On the left, a word cloud is displayed, representing the most common keywords found in the responses. This visualization highlights the frequency and prominence of key terms, offering insights into the main themes and concepts discussed in the answers. On the right, a bar chart shows the distribution of question types, providing an overview of the variety and balance of questions posed during the study. Together, these figures help to further understand the characteristics of the responses and the types of questions that were most prevalent in the dataset
 \subsection{Experimental Settings }
 \label{Evaluation}

 As given in Table \ref{tab:settings}, all models were used in their pre-trained state to ensure a fair comparison across different architectures. Given the diversity of the models employed, specific hyperparameter tuning was not performed for individual models; instead, the focus was on evaluating their inherent capabilities. Each model was assessed under consistent conditions, using a single NVIDIA 24GB GPU to run the experiments, ensuring uniformity in computational resources across the tasks.
\begin{table*}[htb]
    \centering
    \setlength{\tabcolsep}{4pt} 
    \resizebox{\textwidth}{!}{ 
    
    \begin{tabular}{llcl}
        \hline
       \rowcolor[gray]{0.9}
        \textbf{Model} & \textbf{Architecture} & \textbf{Context Length} & \textbf{Evaluation Mode} \\
        \hline
        GPT-4o & closed-source & 2,048 tokens & zeroshot, pre-trained wts \\
       
        GeminiPro1.5 & closed-source & 2,048 tokens & Caption, QA tasks \\
      
        GeminiPro1.0 & Proprietary closed-source & 2,048 tokens & zeroshot, pre-trained wts \\
        
        InstructBLIP & BLIP-2\_Vicuna\_Instruct (Vicuna-7B) & 2,048 tokens & zeroshot, pre-trained wts \\
      
        LLaVA & LLaVA-1.6 (Vicuna-7B) & 2,048 tokens & zeroshot, pre-trained wts \\
        
        Qwen-VL & Qwen-VL-Chat (Qwen-7B) & 2,048 tokens & zeroshot, pre-trained wts \\
      
        CogVLM & CogVLM-chat-v1.1 (Vicuna-7B) & 2,048 tokens & zeroshot, pre-trained wts \\
     
        MiniGPT-4 & MiniGPT-4 (Vicuna-7B) & 2,048 tokens & zeroshot, pre-trained wts \\ \hline
      
    \end{tabular}
    
    }
    \vspace{0.5cm}

    \begin{tabular}{lp{12cm}}
        \hline
     \rowcolor[gray]{0.9}
        \textbf{Evaluation Process} & \textbf{Details} \\
        \hline
        Evaluation Method & Models were evaluated on Waste-Bench tasks, including classification, counting, color recognition, and other categories. GPT-4 evaluated model predictions. \\
        \hline
        Human Verification & Two human evaluators verified model predictions, showing high consistency with GPT-4 evaluations. \\
        
        Error Handling & Default safety mechanisms were employed to prevent out-of-memory errors and ensure stable performance. \\
        \hline
    \end{tabular}
      \caption{Experimental Setup and Model Specifications.}
    \label{tab:settings}
    
\end{table*}
\subsection{Insights}
 \label{Insight}

\begin{figure*}
  \centering
   \includegraphics[width=1\textwidth]{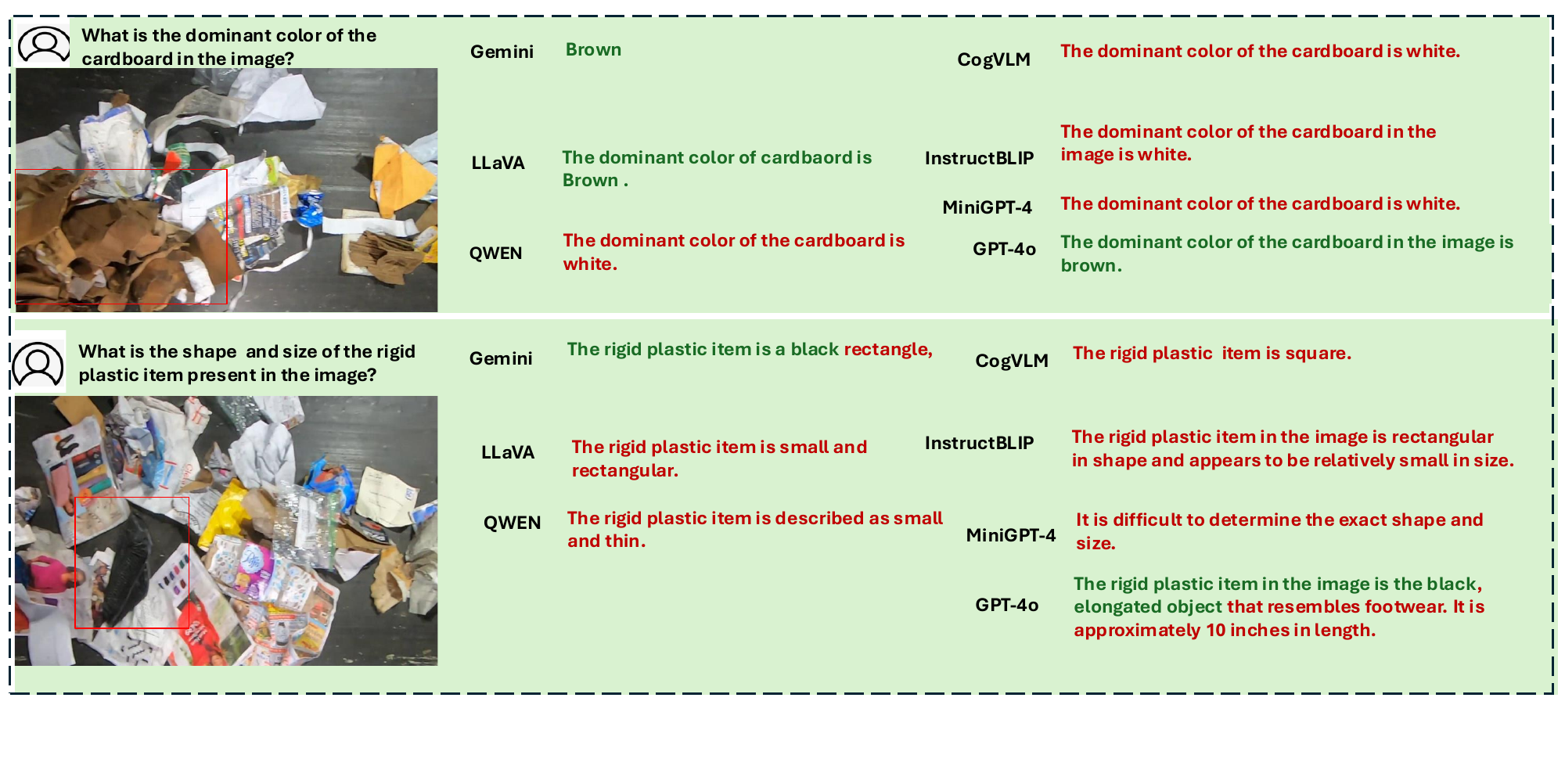}
  \caption{Qualitative results illustrating models struggling with identifying shapes, colors, and recognizing rare classes within cluttered scenes, indicating areas for further investigation and improvement. }
  \label{fig:evaluation22}
  \end{figure*}
\textbf{Recognition and Counting Challenge}:
Models generally struggle with recognizing and classifying objects across all classes in cluttered environments. As illustrated in Figure \ref{fig:evaluation22}
, the models face significant challenges when dealing with complex and cluttered environments, as shown by the incorrect answers highlighted in red. However, we included a case where the models performed better, such as accurately identifying the dominant color in the image, with few models providing the correct answer. This contrast highlights that while models can handle simpler tasks, like recognizing a dominant color in scenarios with clear and singular visual cues, they continue to struggle with more complex tasks that require understanding spatial relationships and object classification in cluttered environments. Including this case emphasizes that while there are areas where models show reasonable performance, significant gaps remain in more challenging real-world scenarios

However, the models struggle significantly when dealing with more complex tasks, like identifying the shape and size of objects or differentiating between similar materials in cluttered environments. Despite clear instructions regarding the presence of only one rigid plastic item, the responses varied widely, highlighting ongoing challenges in spatial reasoning and object recognition. These inconsistencies emphasize that while models can handle basic visual tasks, they falter when faced with more intricate aspects of real-world scenes, such as understanding object relationships or accurately assessing size and material properties\\

\subsection{Challenges with Noise, Enhanced Lighting and Shaded Degradations}:\\
\label{noisy}
While not the main focus of our paper, we further extended our evaluation to assess the models' performance across various degradations. Our experiments revealed that introducing noise, shading, and enhanced lighting conditions in the images exacerbates performance issues in the models, as shown in Table \ref{tab:degradations}. For instance, some models experience a significant drop in accuracy when noise is introduced, highlighting their vulnerability, while others exhibit better noise-handling capabilities. These findings underscore the importance of incorporating environmental factors into future model evaluations. To ensure consistency in our experiments, we applied fixed levels of degradation. Specifically, we used a gradient mask for shading with an initial intensity of 0.7, a Gaussian noise with a sigma value of 7, and a brightness factor of 1.2 for enhanced lighting in the HSV color space. Evaluating these natural degradations is crucial for understanding the robustness of models in real-world scenarios, where ideal conditions are seldom guaranteed. By testing models under these challenging conditions, we are able to identify vulnerabilities and areas for improvement, ensuring that models are better equipped to handle diverse and unpredictable environments. This is also important in considering the performance measure of VLLMs in applications other than waste such as surveillance, autonomous driving, and environmental monitoring, where models need to be resilient to a wide range of environmental factors and disruptions. 
\begin{table}[h]
\centering
\resizebox{\columnwidth}{!}{%

\label{tab:tab-results}
\begin{tabular}{@{}lcccc@{}}
\toprule
\textbf{Model} & \textbf{Normal} & \textbf{Noisy} & \textbf{Enhanced} & \textbf{Shaded} \\ 
\midrule
\textbf{Gpt-4o} & 57.52 & 57.04 & 57.40 & 56.90 \\ \midrule
\textbf{GEMINI} & 49.45 & 48.48 & 48.65 & 48.20 \\ \midrule
\textbf{I.BLIP} & 48.58 & 46.29 & 47.20 & 46.25 \\ \midrule
\textbf{LLaVA} & 47.45 & 47.03 & 46.90 & 46.16 \\ \midrule
\textbf{CogVLM} & 41.58 & 40.15 & 40.50 & 39.73 \\ \midrule
\textbf{Qwen-VL} & 41.30 & 39.40 & 40.58 & 37.09 \\ \midrule
\textbf{MiniGPT4} & 36.40 & 36.21 & 36.90 & 35.20 \\ \bottomrule
\end{tabular}}
\caption{Evaluation results of various Vision Large Language Models (VLLMs) across different degradation scenarios and accuracy metrics.} 

\label{tab:degradations}
\end{table}
\begin{figure*}
  \centering
   \includegraphics[width=1\textwidth]{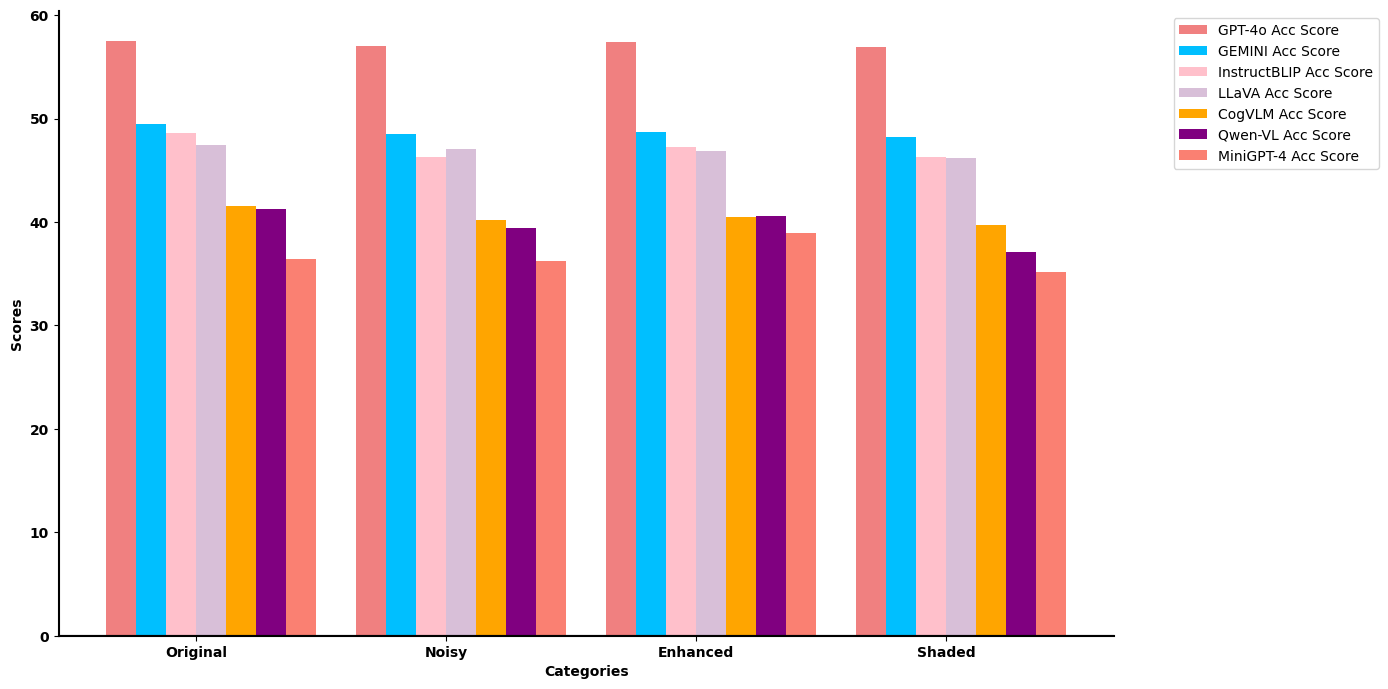}
  \caption{Performance comparison of various Vision Large Language Models (VLLMs) under different degradation scenarios. The chart illustrates how models like GPT-4, GEMINI, InstructBLIP, and others struggle with tasks involving shape recognition, color identification, and classification of rare classes within cluttered scenes, particularly under conditions of noise, enhanced lighting, and shading. This highlights the challenges VLLMs face in maintaining accuracy and robustness when subjected to real-world visual distortions. }
  \label{fig:evaluation2}
  \end{figure*}
\end{document}